# A Correlation Based Feature Representation for First-Person Activity Recognition

**Reza Kahani[1,2], Alireza Talebpour[1,2], and Ahmad Mahmoudi-Aznaveh[2]**

*r.kahani@mail.sbu.ac.ir, talebpour@sbu.ac.ir, a_mahmoudi@sbu.ac.ir*

[1]Department of Computer Science and Engineering, Shahid Beheshti University, Tehran, Iran

[2]Cyberspace Research Institute, Shahid Beheshti University, Tehran, Iran

**Abstract— In this paper, a simple yet efficient activity recognition method for first-person video is introduced. The proposed method is appropriate for representation of high-dimensional features such as those extracted from convolutional neural networks (CNNs). The per-frame (per-segment) extracted features are considered as a set of time series, and inter and intra-time series relations are employed to represent the video descriptors. To find the inter-time relations, the series are grouped and the linear correlation between each pair of groups is calculated. The relations between them can represent the scene dynamics and local motions. The introduced grouping strategy helps to considerably reduce the computational cost. Furthermore, we split the series in temporal direction in order to preserve long term motions and better focus on each local time window. In order to extract the cyclic motion patterns, which can be considered as primary components of various activities, intra-time series correlations are exploited. The representation method results in highly discriminative features which can be linearly classified. The experiments confirm that our method outperforms the state-of-the-art methods on recognizing first-person activities on the two challenging first-person datasets.**

## I. INTRODUCTION

Human action recognition have become an interesting research filed in the recent decade [1-6]. It is because of its numerous applications, such as visual surveillance, entertainment devices, elderly people assistance, human-computer interaction, and video indexing/retrieval. In spite of many efforts conducted on recognition of human activities, it still remains a difficult problem in real-world applications. Intrinsic similarities between different actions give small inter-class variations. On the other hand, there are large intra-class variations caused by camera motion, illumination changes, background clutter, viewpoint changes, irrelevant motions, and various styles/speeds.

The videos taken from an actor's own viewpoint are called first-person videos. Although a lot of research have been conducted on third-person activity recognition, it is not appropriate to directly employ them for first-person videos. It is due to major differences between these two kinds of videos. The main difference is related to the fact that the person wearing the camera is involved in the activity. As a consequence, strong ego-motion is mostly occurred in this kind of videos. It should be noted that in most of the first-person video analysis, a real time response is required; therefore, the computational complexity should be considered more intensively [7].

In recent years, the number of captured videos in first-person viewpoint has rapidly grown due to increasing wearable cameras [8]. A lot of applications have emerged such as life logging, elderly (or blind) people assistance, military applications, and robot vision [9]. However, the approaches specifically proposed for first-person human activity recognition are limited.

On the basis of our preliminary work [10], we introduce a new method to encode CNN features based on the time series correlation. Given a sequence of per-frame feature descriptors, we abstract them into a single vector by computing inter and intra-time series relations. The main motivation is to develop a simple and efficient video encoding which utilize pre-trained CNNs on the relatively small datasets. In the experiments, it is shown that the proposed method outperforms the previous methods on recognizing activities of two public first-person datasets.

The rest of the paper is organized as follows: in section 2, a brief overview of the previous activity recognition methods are provided. In section 3, the proposed method is explained. The implementation details and evaluation of the proposed method are illustrated in section 4.

## II. RELATED WORKS

In order to classify human activities various hand-crafted video features have been proposed [11-19]. Laptev [12] proposed Space Time Interest Point (STIP) which is a 3D extension of 2D Harris detector. Cuboid detector [13] is based on 1D Gabor filters applied on temporal axis. Improved trajectory feature (ITF) [11] is one of the most successful approaches detecting informative regions from videos. It is relied on tracking interest points to obtain constant length motion trajectories. A volume around each trajectory is then described using histogram of oriented gradient (HOG) [14, 15], histogram of optical flow (HOF) [14, 16], and motion boundary histogram (MBH) [16]. SIFT-3D [17], extended SURF [18], and HOG-3D [19] are extension of the baseline approaches to describe videos by considering temporal dimension.

Considering the intrinsic differences between first and third-person videos, several methods have been specifically proposed for first person viewpoint videos [7, 20-25]. In [7], the combination of the local and the global features using a multi-channel kernel is investigated. Furthermore, [7] proposed to explicitly consider temporal structure using a hierarchical structure learning. Narayan *et al.* extend improved trajectory approach [11] by grouping trajectories using a motion pyramidal structure [22]. Kitani *et al.* [21] proposed a framework for ego-centric videos by using a stacked Dirichlet process mixture model to automatically learn a motion codebook and ego-action categories. In another work a set of optical flow based motion features for first-person videos is proposed [23].

Each of these hand-designed features just covers parts of the possible feature space. Therefore, these methods can have effective results only when the videos have a limited diversity. They are not generally appropriate for realistic applications.

The use of deep-learning has been extensively growing in order to recognize human activities in the past few years [26-28]. Simonyan and Zisserman [26] proposed a two-stream ConvNet architecture, containing spatial and temporal networks. The spatial net captures appearance information from individual RGB frames. The input of the temporal net is formed by stacking optical flow fields of consecutive frames. Eventually, a weighted average of the class scores of the spatial and temporal net is used as the final decision.

Wang *et al.* [27] tries to explicitly model long term temporal structure using the temporal segment network. A video is divided into short time snippets, then the final prediction is obtained through a consensus of snippet-level predictions. Feichtenhofer *et al.* extend [26] by combining Two-Stream ConvNet with residual networks [28].

In [29], a deep appearance and motion learning (DAML) is investigated for egocentric videos. For this purpose a deep autoencoder is trained for each appearance and motion network. The output of the networks are finally fused and a non-linear support vector machine is trained to recognize human activities.

It should be noted that all of these methods needs to be trained on a large amount of training data. Despite of this fact, most of the first-person datasets have a limited training videos. In this paper, we try to employ pre-trained deep networks for relatively small egocentric activity datasets without retraining or even fine-tuning the networks. Toward this end, an appropriate feature representation is required.

One of the most popular approaches to represent local features is BoVW [30]. More specifically, BoVW clusters the local descriptors and considers each cluster center as a visual word. Finally, a histogram of the occurrences of each visual word is created for each video. There have been several extensions of this initial idea including kernel-code-book (KCB) that assigns the visual words to a visual vocabulary in a soft manner [31, 32], uses the spatial pyramid [33], and the spatio-temporal pyramid [34] to create local histograms.

Jaakkola and Haussler [35] introduced the baseline of Fisher kernel (FK) encoding, and Perronnin applied it to image categorization [36]. The Fisher kernel encoding can also be considered as another extension of BoVW that captures the first and second order statistics between the feature descriptors and the centers of a trained Gaussian mixture model (GMM). The extension of the Fisher encoding introduced by Perronnin *et al.* [37] is performed by applying a normalization to the Fisher vectors. It has been recently shown that the improved Fisher kernel achieves the best results for many applications such as third-person activity recognition [38].

Most of the first-person action recognition methods have employed the mentioned encoding approaches (i.e. BoVW and IFV) [21, 22, 39]. The main disadvantage of such methods is neglecting the spatial and temporal relations between features while it is very important in first-person videos. In [40], a short time Fourier transform is used to extract the temporal structure of 3D actions on a temporal pyramid. Low-frequency Fourier coefficients are used as the represented feature. Jain *et al.* [41] used the average of CNN features over frames to represent a video sequence; the temporal information is still not effectively considered.

Ryoo *et al.* [42] introduced a first-person specific encoding method by applying different pooling operators over frames and concatenating their results to get a single vector. In addition to employing a temporal pyramid, they proposed to count the number of gradients within the temporal filters ($\Delta 1$) in order to better consider the temporal relations. Piergiovanni *et al.* [43] proposed a model to learn latent sub-events using temporal attention filters (an extension of the spatial attention filters for digit generation [44]). LSTMs (long-short term memory) is used to adjust the temporal filters. The method is evaluated using different input features such as: VGG [45], ITF [11], and TDD features (which integrates ITF and deep convolutional features) [46].

Despite of several research efforts, effectively exploiting the temporal relations is still highly desirable. It is more obvious when the input features have been extracted using a convolutional network with individual RGB frames (e.g. Caffe-net, or VGG-net features).

In this paper, we propose to represent ConvNet features for human action recognition based on inter and intra series relations. The experiments confirm that the temporal relations between features are effectively extracted even without using a temporal ConvNet (without explicitly computation of optical flow between consecutive frames, which is a very expensive operation especially for wearable devices).

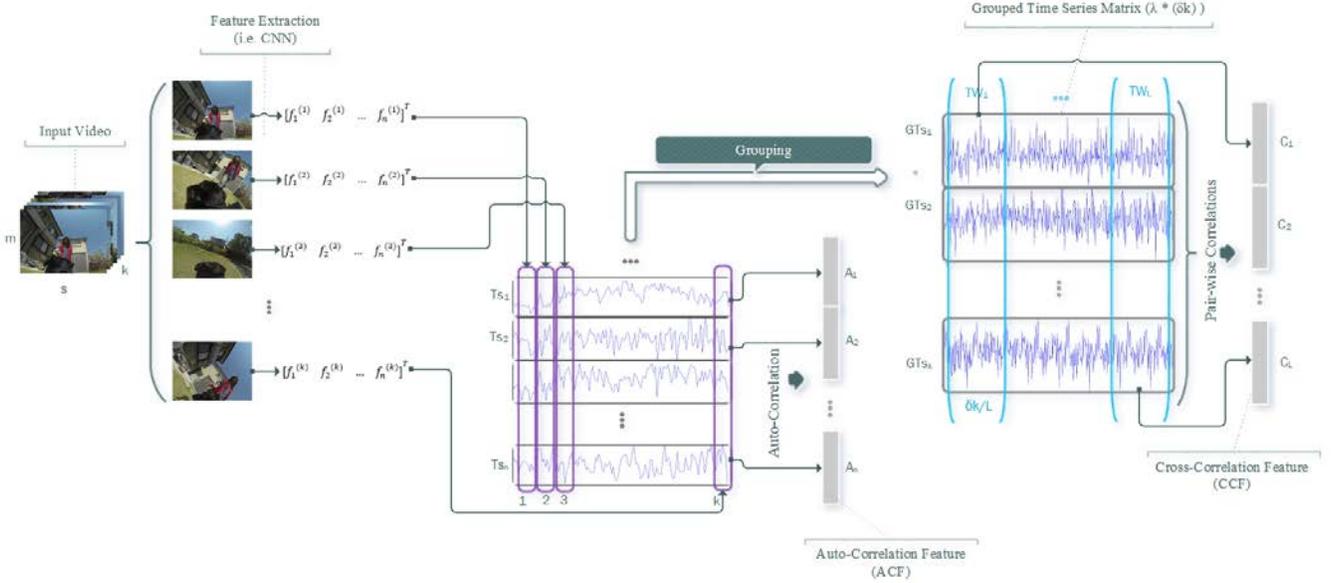

Fig. 1. The proposed video representation Framework. First, features extracted frame by frame using a pre-trained CNN. Then a time series matrix is computed. Finally, the features are represented using a cross and auto correlation on the time series matrix. The concatenation of the represented vectors is used as the final video representation.

## III. CORRELATION BASED FEATURE ENCODING

In this section, the feature extraction step is firstly explained, and then the feature encoding approach based on the time series correlation is proposed.

### A. Feature Extraction

So far, numerous approaches have been proposed to extract features for action recognition such as, low level features that represent a video by describing a number of local interest points, mid-level, and high-level approaches trying to extract high level semantic information. Each of these hand-designed features just covers parts of the possible feature space. Therefore, these methods can have effective results only when the videos have a limited diversity. They are not generally appropriate for realistic applications.

In recent years, deep convolutional neural networks (CNNs) have become an important tool in computer vision tasks. In addition to the significant improvement in image classification [47-50], it has shown effective results in action recognition [26, 46, 51]. However, training a new network is not generally applicable. It is due to the huge number of parameters that must be learnt (tens of millions) in a CNN. In this case, a large set of data is required to train the network while most of the available first-person action datasets are relatively small. A powerful hardware is needed too.

In order to take the benefits of deep learning to extract discriminative features even for small datasets, as well as avoiding the difficulties of training a network, a previously trained network can be used [52]. It is shown that using an image-level CNN features have achieved an impressive result in the first-person activity recognition [42]. In order to use a pre-trained CNN as a feature extractor, the outputs of one of the fully-connected layers (before the last layer) are usually used as a feature vector. In this paper, we employed an image-level CNN as a feature extractor. The method is not limited to special type of CNN features, but in our implementations the output of the first fully connected layer of Caffe [53] and VGG networks [45] have been used. The networks were pre-trained on the ImageNet dataset [54].

When a pre-trained segment-level CNN is used to extract the features from a video sequence, it gives a feature vector for each frame separately. As a consequence, the temporal relations between frames are not explicitly considered. In addition, the final feature dimension is considerably high. On the other hand, due to the variable length of activities, the achieved feature vectors involve a variable-size set of descriptors. As a result, using a feature encoding is necessary to obtain an effective representation for video sequences.

*B. Feature Encoding*

The main idea to encode the features extracted from a video sequence is to capture relations which exist among them. For this purpose, we applied the correlation operator to capture inter and intra relations of the time-series.

Fig. 1 illustrates the overall process of the proposed encoding framework. First, an image-level CNN is employed to extract features of each frame. Then, a time-series matrix is formed by concatenating the feature vectors. After that, the matrix is represented in two ways. Cross-correlation is applied to extract the temporal dynamics while auto-correlation is employed to capture self-similarities. Finally, the achieved features are fused to get the final video representation. The whole procedure will be explained in more details in the following.

First step is the per-frame feature extraction for each video by a pre-trained CNN. Let the feature descriptor obtained for the $t^{th}$ frame denotes as:

$$F^{(t)} = [f_1^{(t)}, f_2^{(t)}, ..., f_n^{(t)}] \tag{1}$$

where $n$ is the number of features at each frame (i.e. $n$ is the number of neurons in a fully-connected layer of the network). Then, a time series matrix ($TS$) is formed by concatenating the frames descriptors:

$$TS = [Ts_1; Ts_2; ...; Ts_n] \tag{2}$$

where

$$Ts_i = [f_i^{(1)}, f_i^{(2)}, ..., f_i^{(k)}] \tag{3}$$

in which $k$ is the number of frames in the video. Each row of the matrix can be considered as a time series.

*1) Inter-Time Series Relation*

There is a little understanding about what spatial features extract from CNNs [55]; however, the relations between them can represent motions and scene dynamics which are more important to capture in the first-person videos [42]. It can be concluded that the temporal relations can be effectively represented using cross-correlation coefficients between the time-series. In order to extract the inter-time series relations a linear cross-correlation is computed between each pair of the time series. The correlation coefficients are used as the encoded vector $C$:

$$C = [r_{Ts_{1,2}}, r_{Ts_{1,3}}, ..., r_{Ts_{n-1,n}}] \tag{4}$$

where,

$$r_{Ts_{a,b}} = \frac{\sum_{i=1}^{k} Ts_a^i Ts_b^i - k(\overline{Ts_a})(\overline{Ts_b})}{(k-1)s_{Ts_a}s_{Ts_b}},$$

$$\forall_{a,b}, a,b \in \{1,2,...,n\} \,\& \,(a<b)$$

(5)

in which $\overline{Ts_a}$ and $s_{Ts_a}$ are the mean and the standard deviation of the $Ts_a$ vector. The length of this vector will be equal to $(n(n\text{-}1))/2$.

**Grouping Strategy:** It should be noted that the vector dimension will not be reasonable, when the parameter *n* is not sufficiently small (e.g. for *n*=4096 the vector *C* will have more than 8 million dimensions). To control the vector length, one way is selecting a limited subset of feature series to compute correlations. However, useful information may be missed in this way. As a result, to control the vector length as well as to avoid discarding features, a grouping strategy is employed. For this purpose, we put each $\delta$ series together as a grouped time series. More specifically, the time series matrix is divided to $\lambda$ horizontal groups. Then, each group of the matrix is vectorized (in a column-wise manner) to form a $\lambda\times(\delta k)$ dimensional grouped time series matrix (*GTS*):

$$GTS = [GTs_1; GTs_2; ...; GTs_n]$$

(6)

where

$$GTs_1 = [TS_1^{(1)}, ..., TS_\delta^{(1)}, ..., TS_1^{(2)}, ..., TS_\delta^{(2)} ... TS_1^{(k)}, ..., TS_\delta^{(k)}],$$

$$\delta = \frac{n}{\lambda}.$$

(7)

The encoded vector *C* for the video is then computed using the correlation coefficients between each pair of the grouped time series:

$$C = [r_{GTs_{1,2}}, r_{GTs_{1,3}}, ..., r_{GTs_{\lambda-1,\lambda}}]$$

(8)

where

$$r_{GTs_{a,b}} = \frac{\sum_{i=1}^{k\delta} GTs_a^i GTs_b^i - k\,\delta(\overline{GTs_a})(\overline{GTs_b})}{(k\,\delta-1)s_{GTs_a}s_{GTs_b}},$$

$$\forall_{a,b}, a,b \in \{1,2,...,\lambda\} \,\& \,(a<b)$$

(9)

in which $\overline{GTs_a}$ and $s_{GTs_a}$ are the mean and the standard deviation of the $GTs_a$ vector. The length of the encoded vector *C* will be $\frac{1}{2}\lambda(\lambda\text{-}1)$. Accordingly, by using the grouping strategy all of the extracted features are used for computing the final representation.

Fig. 2 shows the recognition accuracy of the proposed encoding method under different conditions: with or without grouping strategy. In order to analyze the effect of grouping strategy in controlling the length of encoded feature vectors, three straightforward schemes were also used to select a subset of time series: "First" that selects the first subset of the time series,

"Random" and "Uniform" which select the series randomly, and densely using a uniform stride respectively. This evaluation is performed for various number of series/groups.

The proposed grouping strategy achieves a superior accuracy especially when the number of selected series is small. As it is expected, the grouping strategy leads to a rich feature representation by avoiding to discard the feature series. On the other hand, the grouping strategy improves the classification accuracy than even when all of the series are exploited for feature representation without grouping. It is owing to the fact that the grouping strategy can control the classifier complexity, as an important factor to prevent overfitting, with regard to variation of dataset instances. For instance, by using the proposed grouping with $\delta = 64$, the recognition accuracies improve by 5% on DogCentric dataset. In addition, the final feature dimension is reduced to 2,016D in contrast with 8,386,560D when the grouping is not employed (reduced by 4160X). The datasets are introduced in more detail in section IV.A. As a consequence, by choosing a suitable value for $\delta$ the method can utilize all feature series with an impressive lower dimension as well as improving the final accuracy.

The proposed grouping strategy is simple yet efficient. It does not impose extra overhead to the overall procedure. The major superiority of our grouping strategy over the conventional dimension reduction methods is its ability to be applied without a training phase; more specifically, the grouping strategy can be employed independent of the other sequences. The common dimension reduction methods require a large data with high diversity in order to obtain an effective model. It is in contrast with the fact that most of the available first-person datasets are relatively small. Furthermore, training phase demands a high computational cost which may be infeasible for a representative training set. In the test phase, computation time of the correlation based encoded vector is also extremely reduced with the grouping strategy (more than 42X) regardless of the offline training time. In summary, unlike the common dimension reduction methods, the proposed grouping strategy can be applied effectively.

**Temporal Partitioning**: In order to effectively represent the long-term movement, the series length should not be very large. To control the series length and focus on each local time, we employ a number of non-overlapping uniform time intervals:

$$TW_i = [(i-1) \times D + 1, i \times D], \; i = 1, 2, ..., L$$
$$\text{where } D = \frac{K\delta}{L} \tag{10}$$

$TW_i$ is the $i$-th time interval and $L$ indicates the number of intervals. In other words, we divide the grouped time series matrix to $L$ vertical parts and encode each part separately. Finally, the local encoded vectors are concatenated to achieve the Cross Correlation Feature vector:

$$CCF = [C_1, C_2, ..., C_L] \tag{11}$$

consequently the encoded vector $CCF$ has $L(\frac{1}{2}\lambda(\lambda-1))$ dimensions. It should be noted that the incorporated temporal partitioning leads to track sequence variations over time. Furthermore, it also can help to avoid missing local motions.

*2) Intra-Time Series Relation*

In order to consider temporal information more precisely and extract repeating patterns, we measure self-similarities for each feature series. Our motivation is to effectively capture the temporal self-similarities which arise from the fact that many parts of a video sequence are similar.

For this purpose, the time series matrix is first formed; then, sample autocorrelation with $\gamma$ lags and a constant stride is computed for each feature series (each row of the matrix, i.e. 4096). Finally, these correlation coefficients are concatenated to obtain the Auto

Correlation Feature vector (*ACF*). Length of this vector is $n\gamma$. It is notable that the parameter $\gamma$ is dependent on factors such as frame rate, sequences duration, and the actions execution speed.

The final Time series Correlation Feature vector (*TCF*) is composed of concatenation of the vector *CCF* and the vector *ACF* (i.e. *TCF*=[*CCF*, *ACF*]) and has $(n\gamma) + (L(\frac{1}{2}\lambda(\lambda-1)))$ dimensions. The experiments demonstrate that the features represented using the cross-correlation and auto-correlation are complement with each other.

## IV. EXPERIMENTS

In this section, we first introduce the datasets. Then the experimental setup and the parameters setting are explained. Next the proposed method is compared with the state of the art on two challenging first-person dataset: DogCentric, and UEC-Park. Finally, an experimental analysis is performed in order to provide a more comprehensive evaluation.

In all our experiments, we randomly selected half of the video sequences of each activity for training and used the rest of them for evaluation. (The number of training videos for each class will be less than or equal to the number of test sequences) We repeated this random data splitting for 100 times and reported the mean accuracies. A one-vs-rest linear SVM is used as the classifier of the proposed method in all experiments. The regularization parameter (C) is set to 1000.

### A. Datasets

**Dog-Centric[1]:** The DogCentric [20] is a very challenging dataset composed of first-person animal videos. It consists of 209 video sequences of 10 activities performed by the dogs wearing a camera. The dataset contains two types of activities (i.e. animal ego-action and human-animal interaction). It should be noted that most of the video sequences contain a heavy amount of ego-motion. As it is shown in Table I, videos are not uniformly distributed in all classes, as well as the videos length varies widely (between 30 to 650 frames). The Fig. 3-left shows one frame of two different activities from this dataset.

**UEC-Park[2]:** the UEC-Park dataset [21] is a challenging 25 minute workout video sequence that captured in a first-person point of view. We segmented this sequence at the rate of one segment every two seconds. Moreover, the frame rate is halved and each frame is down-sampled by factor of two. The video distribution in classes is very unbalanced (i.e. between 1 to 119 clips for each class). Two snapshots of Park activities are shown in Fig. 3-right. Full details about each dataset are shown in Table I

TABLE I
THE PROPERTIES OF THE EMPLOYED DATASETS

| Dataset | DogCentric | UEC-Park |
|---|---|---|
| Class | 10 | 29 |
| Video | 209 | 766 |
| Resolution | 320×240 | 848×480 |
| FPS | 48 | 60 |
| Clips | 10-27 | 1-119 |
| Frames | 30-650 | 120 |
| Year | 2014 | 2011 |

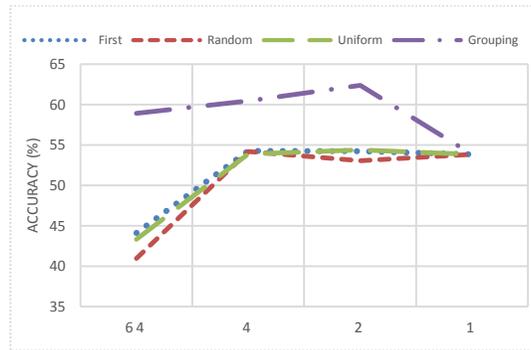

Fig. 2 The effect of Grouping Strategy on DogCentric dataset

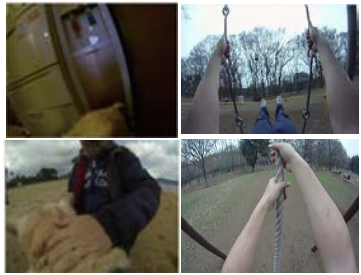

Fig. 3. Examples of video frames: (Left) Dog-Centric, (Right) UEC-Park.

### B. Experimental Setup

**Parameters Evaluation**: In order to find the best parameters, different settings are investigated. For the sake of concise presentation, the analysis of various parameter settings for the DogCentric dataset is only described.

In all experiments, we fixed $\lambda$ to 64 ($\delta = n/\lambda$) for the deep network features. Since it has shown a better performance for both DogCentric and UEC-Park datasets with a very compact feature dimension, either using Caffe-Net or VGG-Net as feature extractor.

As it is mentioned before, the efficient number of lags ($\gamma$) for calculating the auto-correlation is related to factors such as frame rate, sequences duration, and the actions execution speed. On the other hand, the number of non-overlapping windows, $L$, help to preserve the long-term movements. In order to find the best settings for the parameters $L$ and $\gamma$, we explore different values for each on the DogCentric dataset. As is shown in Fig. 4, for each value of $L$, we vary $\gamma$ from 1 to 7 and the recognition rate are illustrated. As the results show, $L=16$ and $\gamma=6$ gives the best performance for the DogCentric dataset. As Fig. 4 demonstrates, for each $L$ a large number of lags leads to a redundant feature set implying that overfitting is more likely to occur. In the case of TDD features, we fixed $L$ to one in all experiments.

### C. Results

In this section, the proposed method is first compared with the previous representation methods using the same experimental conditions. After that, the proposed method is also compared with state-of-the-art approaches in terms of recognition accuracy. All the experiments performed on two challenging first-person datasets (i.e. DogCentric and UEC-Park). In the following, the results for each dataset is reported separately.

In the experiments, the image-level CNN features extracted from the output of Caffe-Net [53] and VGG-Net [45] models. In the case of TDD feature [46], we use the combination (concatenation) of the conv4 output descriptor of spatial-net and the conv4 output descriptor of temporal-net as the input of the proposed method.

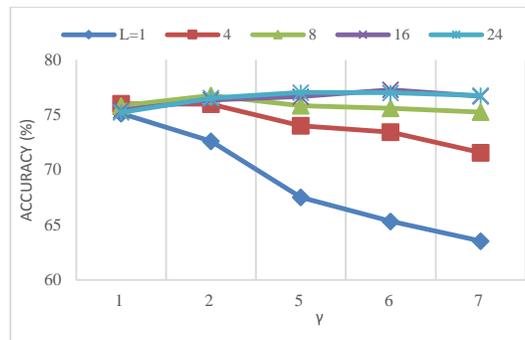

Fig. 4. The effect of the number of non-overlapping time windows ($L$) and the number of lags ($\gamma$) on the overall accuracy for the DogCentric dataset

TABLE II

COMPARISON OF THE REPRESENTATION METHODS ON THE DOG-CENTRIC DATASET

|  | Method | Accuracy (%) |
|---|---|---|
| **Caffe** | FTP [40] | 64.87 |
|  | IFV | 54.53 |
|  | PoT [42] | 63.94 |
|  | **TCF (Proposed)** | **72.19** |
| **VGG** | Sum Pooling [42] | 60.00 |
|  | Simonyan [45] | 59.90 |
|  | Temporal Attention Filters [43] | 65.50 |
|  | **TCF (Proposed)** | **77.79** |

**DogCentric**: Table II shows the mean recognition accuracy of the proposed method and the other representation approaches after 100 repetitions. Experiments are reported using two image-level CNN features (i.e. Caffe-Net and VGG-Net).

It can be observed that the proposed method has significantly better accuracy than IFV (about 17.6%). We believe it is because of the fact that IFV miss the temporal relations between features. Furthermore, slight changes are missed in the quantization step especially when the number of frames is not sufficiently higher than the feature dimension.

In contrast to the best previous representation results, the proposed method has a significant improvement (i.e. about 8.25% using Caffe-Net feature and 12.29% using VGG-Net feature) in recognition accuracy. It can be concluded that our method can extract temporal relations more effectively. In addition, the proposed method explicitly extracts the cyclic patterns from each video. Furthermore, the results confirm that the temporal relations between features are effectively extracted even without using a temporal ConvNet (without explicitly computation of optical flow between consecutive frames, which is a very expensive operation especially for wearable devices).

The proposed approach with the state-of-the-art recognition methods are also compared. In this experiment, we also employ another method as a feature extractor which called TDD [46]. (TDD is based on a combination of Deep image-level CNN and Trajectory feature). The result confirm that the proposed method can improve the recognition accuracy even using TDD (i.e. about 3.28%). The combination of two features (VGG and TDD) further improves the recognition rate. To the best of our knowledge, this is the best result on this dataset. Confusion matrix for the proposed method is shown in Fig. 5.

TABLE III

COMPARISON OF THE PROPOSED METHODS WITH THE STATE-OF-THE-ART METHODS ON THE DOGCENTRIC DATASET IN TERM OF RECOGNITION ACCURACY. THE

FEATURE TYPES FOR SOME METHODS HAVE BEEN WRITTEN IN THE BRACKETS.

| Method | Accuracy (%) |
|---|---|
| Inria ITF (with IFV) [11] | 67.58 |
| Iwashita [20] | 60.50 |
| Amsterdam [41] | 69.23 |
| PoT [MBH, HOF, OverFeat, Caffe]+ ITF [42] | 74.47 |
| PoT [42] | 73.00 |
| PoT (CaffeNet + VGG + C3D + HOF) [56] | 70.79 |
| RMF [23] | 61.00 |
| TDD [46] | 76.60 |
| Temporal Attention Filters (temporal filters) [43] | 79.60 |
| Temporal Attention Filters (temporal filters + LSTM) [43] | 81.40 |
| VGG [45] | 59.90 |
| SimpleMKL [25] | 64.90 |
| CDN [57] | 77.20 |
| Trajectory VRTD [24] | 69.60 |
| **TSC (Proposed) [VGG]** | **77.79** |
| **TSC (Proposed) [Caffe] + ITF** | **77.49** |
| **TSC (Proposed) [TDD]** | **79.88** |
| **TSC (Proposed) [VGG, TDD]** | **82.24** |

**UEC-Park:** In these dataset, the length of videos is relatively short. Moreover, the activities are not aligned in the videos. Table IV show the final recognition accuracies of the feature representation approaches on the Park dataset. It is clear that the proposed encoding method could successfully encode image-level features to a global vector.

We also compared the method with the state-of-the-art recognition methods on the Park dataset. As the result is shown in Table V, the recognition rate is comparable with the best previous result.

|  | Ba | Ca | Dr | Fe | Le | Ri | Pe | Sh | Sn | Wa |
|---|---|---|---|---|---|---|---|---|---|---|
| Ball | 83.3 | | 0.4 | 0.4 | | | | 13.4 | 2.4 | |
| Car | | 97.8 | 0.8 | 0.1 | | 0.2 | 0.9 | | | 0.2 |
| Drink | 6.0 | 3.6 | 64.4 | 3.4 | 0.4 | | 0.6 | | 21.6 | |
| Feed | 2.8 | 0.1 | 1.1 | 78.1 | 1.2 | 0.6 | 10.3 | 0.2 | 0.6 | 5.2 |
| LookLeft | 0.6 | 0.3 | | 6.7 | 63.2 | 4.7 | 0.6 | 1.9 | 2.1 | 19.8 |
| LookRight | 0.4 | 1.8 | | 3.7 | 5.0 | 51.2 | 4.1 | 0.2 | 2.4 | 31.1 |
| Pet | | 2.2 | | 4.6 | 0.7 | | 92.4 | | 0.2 | |
| Shake | 7.2 | | | | | | 4.0 | 88.8 | | |
| Sniff | 0.8 | | 0.2 | 0.3 | | | | 0.6 | 95.3 | 2.9 |
| Walk | 0.2 | 0.1 | 0.1 | 0.4 | 4.8 | 5.1 | | 0.2 | 3.1 | 86.0 |
| Total Accuracy | | | | | | | | 82.24 | | |

Fig. 5  Confusion matrix for the proposed method on the dog-centric Dataset.

*D. The effect of inter and intra time relations*

In this experiment, the recognition ability of the proposed method is evaluated while only one of the two encoded feature is used. The aim is to show the contribution of each representation scheme (Inter/Intra) on the overall recognition accuracy. Fig. 6 illustrates the effect of each part of the representation scheme on the Dog-Centric and Park datasets.

This experiment confirms that jointly employing two types of representation will benefit the overall recognition. In other words, the intra-time relations could impressively complement cross-correlation based features to enhance the final recognition accuracy.

TABLE IV

MEAN RECOGNITION ACCURACY FOR REPRESENTATION METHODS ON THE UEC-PARK DATASET

|  | Method | Accuracy (%) |
|---|---|---|
| **Caffe** | BoVW | 64.64 |
|  | IFV | 67.79 |
|  | PoT [42] | ~ 73.00 |
|  | FTP [40] | 70.63 |
|  | **TSC (Proposed)** | **73.06** |
| **VGG** | BoVW | 65.25 |
|  | IFV | 67.86 |
|  | PoT [42] | _ |
|  | **TSC (Proposed)** | **73.18** |

TABLE V

COMPARISON OF THE PROPOSED METHOD WITH THE STATE OF THE ARTS ON THE UEC-PARK DATASET IN TERM OF RECOGNITION ACCURACY

| Method | Accuracy (%) |
|---|---|
| STIP (with IFV) [12] | 69.13 |
| Cuboid (with IFV) [13] | 72.33 |
| IFV [MBH, HOF, OverFeat, Caffe] | 76.40 |
| Inria ITF (with IFV) [11] | 76.62 |
| ITF + CNN [41] | 75.74 |
| PoT + ITF [42] | 79.49 |
| DAML [29] | 77.55 |
| CDN (ResNet-50) [57] | 78.70 |
| **TSC (Proposed) [Caffe, TDD] + ITF** | **79.69** |
| **TSC (Proposed) [TDD] + ITF** | **79.26** |

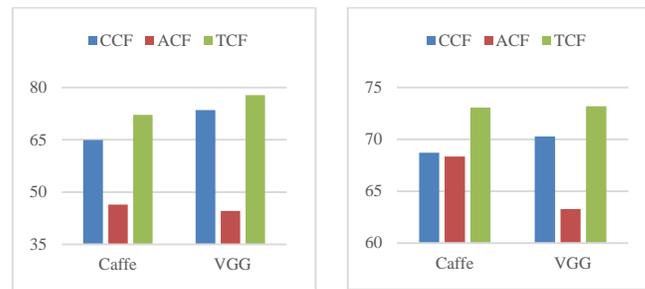

Fig. 6 the effect of inter and intra time relations. (Left) DogCentric, (right) UEC-Park

## V. Conclusion

In this paper, we proposed an activity recognition method which can effectively capture the temporal relations among per-segment features. The method is relied on capturing the inter and intra-time series relations using linear correlations. The inter-time relations can effectively represent the motion dynamics, besides the intra-time relations capture the temporal self-similarities. In the first-person video analysis the computational costs are more important due to the computing power limitations and the fast response requirement. The experimental results confirm that even without explicitly employing a temporal ConvNet, the temporal relations between features are effectively extracted (i.e. without explicitly computation of optical flow between consecutive frames). In order to control the classifier complexity, a grouping strategy is also introduced. The experiments show that our method outperforms the state-of-the-art on the two challenging first-person datasets. In the future, we aim to exploit the proposed representation method for other video analysis tasks such as scene classification and video retrieval.